 \documentclass[pmlr,twocolumn]{jmlr} 

\usepackage{booktabs}
\usepackage[load-configurations=version-1]{siunitx} 

 \usepackage{color}
\usepackage[all]{xy}

\usepackage{wrapfig}
\usepackage{epsf}
\usepackage{enumitem}
\usepackage[switch]{lineno}  %

\theorembodyfont{\upshape}
\theoremheaderfont{\scshape}
\theorempostheader{:}
\theoremsep{\newline}

\newcommand{\bfX}{\boldsymbol{X}}

\newcommand{\ci}{\perp\!\!\!\perp}

\newcommand{\bfC}{\boldsymbol{C}}

\newcommand{\bfU}{\boldsymbol{U}}

\newcommand{\bfS}{\boldsymbol{S}}
\newcommand{\bfV}{\boldsymbol{V}}

\newcommand{\bfs}{\boldsymbol{s}}
\newcommand{\bfv}{\boldsymbol{v}}

\newcommand{\bfBeta}{\boldsymbol{\beta}}
\newcommand{\bfSigma}{\boldsymbol{\Sigma}}

\newcommand{\bfzero}{\boldsymbol{0}}

\newcommand{\beginsupplement}{%
        \setcounter{table}{0}
        \renewcommand{\thetable}{S\arabic{table}}%
        \setcounter{figure}{0}
        \renewcommand{\thefigure}{S\arabic{figure}}%
     }

\makeatletter
\let\old@biblabel\@biblabel
\def\@biblabel#1{\kern\bibindent}
\let\old@bibitem\bibitem
\def\bibitem#1{\old@bibitem{#1}\leavevmode\kern-\bibindent}
\makeatother

\jmlrvolume{ML4H Extended Abstract Arxiv Index}
\jmlryear{2020}
\jmlrsubmitted{2020}
\jmlrpublished{}
\jmlrworkshop{Machine Learning for Health (ML4H) 2020} 

\title[Stable predictions for anticausal ML health applications]{Stable predictions for health related anticausal prediction tasks affected by selection biases: the need to deconfound the test set features}

\author{%
\Name{Elias Chaibub Neto} \Email{elias.chaibub.neto@sagebase.org}\\
\addr Sage Bionetworks
\AND
\Name{Phil Snyder, Solveig K. Sieberts, Larsson Omberg} \Email{}\\
\addr Sage Bionetworks
}

\editor{}

\begin{document}

\maketitle

\begin{abstract}
In health related machine learning applications, the training data often corresponds to a non-representative sample from the target populations where the learners will be deployed. In anticausal prediction tasks, selection biases often make the associations between confounders and the outcome variable unstable across different target environments. As a consequence, the predictions from confounded learners are often unstable, and might fail to generalize in shifted test environments. Stable prediction approaches aim to solve this problem by producing predictions that are stable across unknown test environments. These approaches, however, are sometimes applied to the training data alone with the hope that training an unconfounded model will be enough to generate stable predictions in shifted test sets. Here, we show that this is insufficient, and that improved stability can be achieved by deconfounding the test set features as well. We illustrate these observations using both synthetic data and real world data from a mobile health study.
\vskip -0.1in
\end{abstract}

\section{Introduction}
\label{sec:intro}
\vskip -0.1in
A standard assumption in supervised machine learning (ML) is that the training and test sets are independent and identically distributed. In practice, however, this assumption is often violated, and dataset shifts (Quinonero-Candella et al 2009) are commonly observed in the real world. At the same time, ML models are often capable of leveraging subtle statistical associations between the input ($\bfX$) and outcome ($Y$) variables in the training data, including spurious associations generated by confounders ($\bfC$) and other sources of biases in the data. As a consequence, predictions from confounded learners are often unstable across shifted test sets, and can fail to generalize.

In anticausal prediction tasks (i.e., tasks where the outcome is a cause of the input variables) dataset shifts in the joint distribution of the confounders and outcome variable, $P(\bfC, Y)$, are often caused by selection biases. As described in more detail in the Background section, selection biases (Heckman 1979; Bareinboim and Pearl 2012) occur when certain subpopulations are under-represented (or over-represented) in the training data relative to the test data, so that $P(\bfC, Y)$ is different in these datasets. (See Supplementary Section 7, for an illustrative hypothetical example.)

While simple approaches such as matching and inverse probability weighting can be used to neutralize these issues in situations where the joint distribution of $\bfC$ and $Y$ in the target population is known (see Supplementary Section 8 for an example), here we focus on the case where the test set can be shifted in unknown ways w.r.t. $P(\bfC, Y)$. This more challenging setting requires more sophisticated adjustment methods, which we review and discuss in Supplementary Section 9. These approaches, however, are sometimes applied to the training data alone with the hope that training an unconfounded model will be enough to generate stable predictions in shifted test sets. Here, we show that this is insufficient, and that deconfounding both the training and test set features can produce more stable predictions. We illustrate this point using the causality-aware approach (Chaibub Neto 2020a) which is able to leverage unlabeled test set data in order to adjust the test set features, even when the shifts in $P(\bfC, Y)$ are unknown.
\vskip -0.1in
\section{Background}
\vskip -0.1in
Throughout the paper we let $\bfX$, $\bfC$, and $Y$ represent the features, confounders, and outcome variables, respectively. The confounded anticausal prediction task influenced by selection bias is described by the causal graph in Figure \ref{fig:anticausal.task}, where the auxiliary
\begin{wrapfigure}{r}{0.19\textwidth}
\vskip -0.3in
$$
\xymatrix@-1.2pc{
 & *+[F-:<10pt>]{\bfC} \ar[dl] \ar[dr] \ar[r] & *+[F]{S} \\
*+[F-:<10pt>]{\bfX}  & & *+[F-:<10pt>]{Y} \ar[ll] \ar[u]}
$$
\vskip -0.1in
\caption{}
\vskip -0.1in
\label{fig:anticausal.task}
\end{wrapfigure}
variable $S$ indicates the presence of a selection mechanism contributing to the association between $\bfC$ and $Y$ (see Suppl. Section 7 for further details). We assume that the causal effect of $\bfC$ on $Y$ is stable, and that the dataset shifts in $P(\bfC, Y)$ are generated by selection mechanisms. We also assume that the causal effects of $\bfC$ and $Y$ on $\bfX$ are stable, so that $P(\bfX \mid \bfC, Y)$ does not change between training and test sets.

In the particular context of linear models, these stable causal effects are represented by stable path coefficients (Wright 1934) in the linear structural equations describing the anticausal prediction task. As described in Chaibub Neto (2020a), the causality-aware approach is implemented by generating counterfactual data that no longer contains the spurious associations generated by the confounders, and only retains the associations generated by the causal effects of the output on the features. For example, in the particular case of a single feature and single confounder we have that the linear structural equation for the feature is given by $X = \beta_{XC} \, C + \beta_{XY} \, Y + U_X$, while the causality-aware feature is estimated by, $\hat{X}_{tr}^\ast = \hat{\beta}^{tr}_{XY} \, Y_{tr} + \hat{U}_X = X_{tr} - \hat{\beta}^{tr}_{XC} \, C_{tr}$ in the training set, and by $\hat{X}_{ts}^\ast = X_{ts} - \hat{\beta}^{tr}_{XC} \, C_{ts}$ in the test set\footnote{Here, we use the subscripts (superscripts) $``tr"$ and ``$ts$" to represent the training and test sets, respectively.}. (See Chaibub Neto (2020a) for the general case.) Note that under the assumptions that the causal effects are stable and that the training set is large, we have that $\hat{\beta}^{tr}_{XC}$ converges to $\beta_{XC}$, so that $\hat{X}_{ts}^\ast$ converges to,
\vspace{-0.1cm}
{\small
\begin{equation}
X_{ts}^\ast = X_{ts} - \beta_{XC} \, C_{ts} = \beta_{XY} \, Y_{ts} + U_X~.
\label{eq:ca.test.feature}
\end{equation}}
\vskip -0.2in
In the next section we present our main contribution in the particular context of linear models\footnote{As described in Chaibub Neto (2020b), the causality-aware approach can also be used to deconfound feature representations learned by deep neural network (DNN) models. This point is illustrated in Section 5 with a real data example.}.
\vspace{-0.5cm}
\section{On the need to deconfound the test set features}
\vskip -0.1in

While it might seen intuitive that training a learner on unconfounded data will prevent it from learning the confounding signal and, therefore, will lead to more stable predictions in shifted target populations\footnote{Examples of approaches that only adjust the training data include pre-processing techniques to reduce discrimination in ML (Calders, Kamiran, and Pechenizkiy 2009; Kamiran and Calders 2012). See also Supplementary Section 9.1.}, here we show that this is insufficient and better stability can be achieved by deconfounding the test set features as well.


We illustrate this point using a toy linear model example (our result, nonetheless, holds for more general linear models, as described in Supplementary Section 10). Assuming (without loss of generality) that the data has been centered, we consider the causal graph $\xymatrix@-1.0pc{C \ar[r] \ar@/^0.5pc/[rr] & X & Y \ar[l]}$ where $C = U_C$, $Y = \beta_{YC} \, C + U_Y$, and $X = \beta_{XY} \, Y + \beta_{XC} \, C + U_X$, with $E[U_V] = 0$, $Var(U_V) = \sigma^2_V$, for $V = \{C, Y, X\}$.
The goal is to predict the outcome $Y$ using the feature $X$. Let $\hat{Y} = X_{ts} \hat{\beta}_{tr}$ represent the test set prediction from a linear regression model, where $\hat{\beta}_{tr}$ represents the coefficient estimated with the training data, and $X_{ts}$ represents the test set feature. By definition the expected MSE is given by,
\vskip -0.2in
{\small
\begin{align}
E&[(Y_{ts} - \hat{Y})^2] = E[Y^2_{ts}] + E[\hat{Y}^2] - 2 E[\hat{Y} Y_{ts}] \nonumber \\
&= Var[Y_{ts}] + E[\hat{Y}^2] - 2 Cov(\hat{Y}, Y_{ts}) \label{eq:expected.mse} \\
&= Var[Y_{ts}] + \hat{\beta}_{tr}^2 Var[X_{ts}] - 2 \hat{\beta}_{tr} Cov(X_{ts}, Y_{ts}), \nonumber
\end{align}}
\vskip -0.2in
\noindent where the expectation is w.r.t. the test set (so that $\hat{\beta}_{tr}$ represents a fixed constant).

For any approach which does not process the test set features we have that,
\vskip -0.2in
{\small
\begin{align*}
Var&(X_{ts}) = Var(\beta_{XY} \, Y_{ts} + \beta_{XC} \, C + U_X) \\
&= \sigma^2_X + \beta_{XY}^2 \, Var(Y_{ts}) + \beta_{XC}^2 \, Var(C_{ts}) \, + \\
&\;\;\;\; + \, 2 \, \beta_{XY} \, \beta_{XC} \, Cov(Y_{ts}, C_{ts})~, \\
Cov&(X_{ts}, Y_{ts}) = Cov(\beta_{XY} \, Y_{ts} + \beta_{XC} \, C + U_X, Y_{ts}) \\
&= \beta_{XY} \, Var(Y_{ts}) + \beta_{XC} \, Cov(Y_{ts}, C_{ts})~,
\end{align*}}
\vskip -0.3in
\noindent showing that both $Var(X_{ts})$ and $Cov(X_{ts}, Y_{ts})$ depend on $Cov(Y_{ts}, C_{ts})$ (so that the $E[MSE]$ will be unstable under dataset shifts of the association between the confounder and the outcome variable across the test sets). Note that this is true even when we apply a confounding adjustment to the training set (a situation where the $\hat{\beta}_{tr}$ estimate is not influenced by the spurious associations generated by the confounder).

On the other hand, from eq. (\ref{eq:ca.test.feature}) we have that for the causality-aware approach,
\vskip -0.2in
{\small
\begin{align*}
Var(X_{ts}^\ast) &= Var(\beta_{XY} \, Y_{ts} + U_X) \\
&= \sigma^2_X + \beta_{XY}^2 \, Var(Y_{ts})~, \\
Cov(X_{ts}^\ast, Y_{ts}) &= Cov(\beta_{XY} \, Y_{ts} + U_X, Y_{ts}) \\
&= \beta_{XY} \, Var(Y_{ts})~,
\end{align*}}
\vskip -0.3in
\noindent does not depend on $Cov(Y_{ts}, C_{ts})$, and will be stable w.r.t. this particular type of dataset shift (although it will be still influenced by dataset shifts on $Var(Y_{ts})$\footnote{Note that the predictions based on the unadjusted test feature will also depend on $Var(Y_{ts})$ (in addition to $Var(C_{ts})$ and $Cov(Y_{ts}, C_{ts})$). Note that, as shown by eq. (\ref{eq:expected.mse}), the expected MSE of any prediction will depend on $Var(Y_{ts})$.}).
\vspace{-0.5cm}
\section{Synthetic data experiments}
\vskip -0.1in
Here, we illustrate the better stability of the causality-aware approach under dataset shifts caused by selection biases in two synthetic data experiments. The first is a regression task where predictive performance is evaluated with MSE. This experiment, compares three adjustment approaches: the causality-aware, where we adjust both the training and test set features; a ``poor man's" version of the causality-aware, where we only adjust the training set features; and the ``no adjustment" approach, where we evaluate the stability in the fully confounded data. The second experiment is a classification task where we evaluate predictive performance using AUROC. In addition to the three adjustment approaches described above we also evaluate under-sampling (by matching) applied to the training data alone in this experiment.

In each experiment we evaluate the performance of the trained model across 9 distinct test sets showing increasing amounts of dataset shift in $P(\bfC, Y)$ (generated by varying $Cov(C_{ts}, Y_{ts})$, $Var(C_{ts})$ and $Var(Y_{ts})$, as described in Supplementary Section 11). The causal effects $\bfBeta_{XY}$, $\bfBeta_{XC}$, and $\beta_{YC}$ were the same across the training and test sets to guarantee that $P(\bfX \mid \bfC, Y)$ was stable.

\begin{figure}[!h]
\vskip -0.1in
\centerline{\includegraphics[width=2.5in]{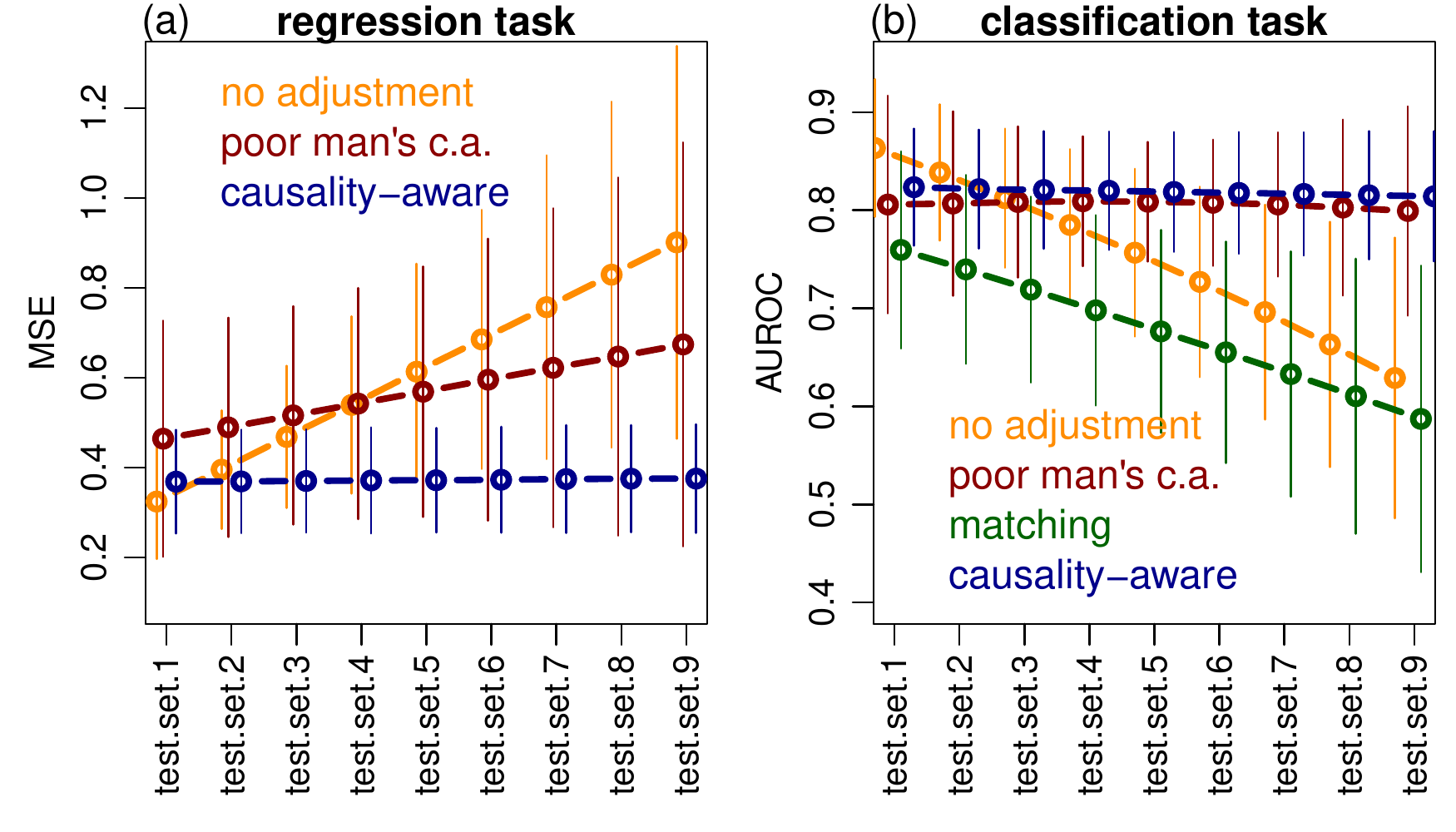}}
\vskip -0.2in
\caption{Synthetic data experiments.}
\label{fig:synthetic.data.stability}
\vskip -0.1in
\end{figure}

Figure \ref{fig:synthetic.data.stability} reports the results (based on 1000 replications) and illustrates the better stability of the causality-aware approach in both experiments. (In both plots the dots represent the mean across the 1000 replications, while the vertical bars represent one standard deviation around the mean.)
\vspace{-0.5cm}
\section{Real data experiments}
\vskip -0.1in
Here, we illustrate the better stability of the causality-aware approach using logistic regression models trained on feature representations learned by deep neural network (DNN) models. As pointed in Chaibub Neto (2020b) the causality-aware approach can still be employed to deconfound feature representations learned by DNNs\footnote{The key idea is that by training a highly accurate DNN using softmax activation at the classification layer, we have that, by construction, the feature representation learned by the last layer prior to the output layer will fit well a logistic regression model (since the softmax activation used to classify the outputs of the DNN is essentially performing logistic regression classification).}. In our experiments we adopted the features learned by a top performing deep learning team in the Parkinson's Disease Digital Biomarker Dream Challenge (Sieberts et al. 2020), where the goal is to build classifiers of disease status (i.e., Parkinson's disease (PD) vs non-PD) using accelerometer data. We split the data into a training set and 2 test sets: one having the same joint distribution of PD labels and (discretized) age confounder as the training set, denoted the ``no shift" test set; and the other having a flipped association between the PD labels and the age confounder, denoted the ``shifted" test set.

\begin{wrapfigure}{r}{0.22\textwidth}
\vskip -0.5in
\centerline{\includegraphics[width=1.3in]{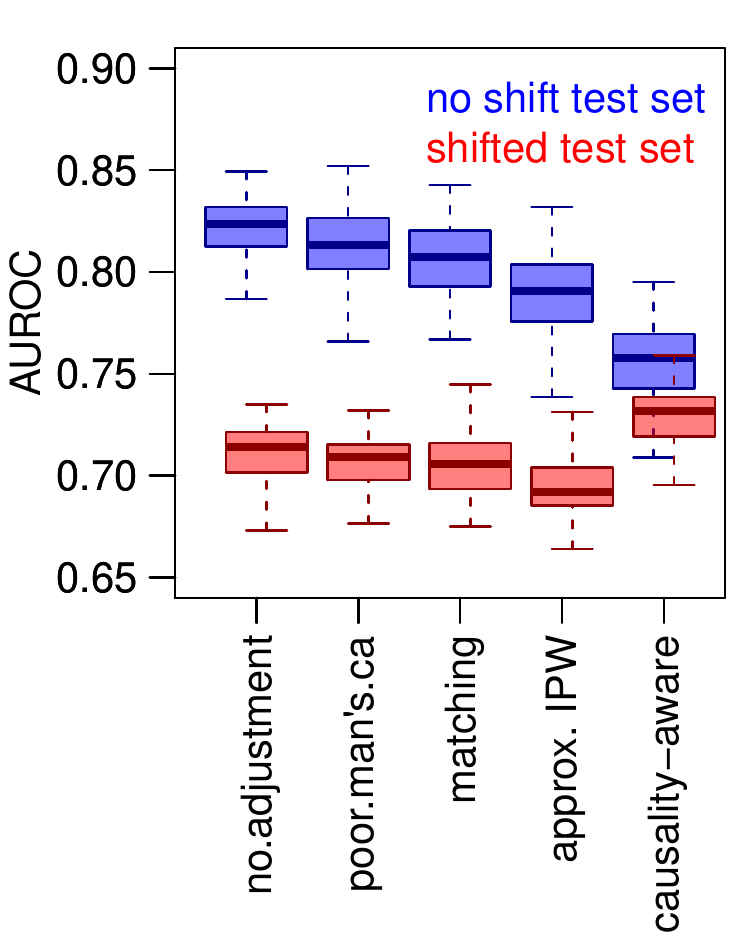}}
\vskip -0.1in
\caption{Results.}
\label{fig:mpower.aucs.stability}
\vskip -0.1in
\end{wrapfigure}

In this illustration we compare the causality-aware adjustment against matching, approximate IPW, ``poor man's" causality-aware, and the ``no adjustment" approaches. Figure \ref{fig:mpower.aucs.stability} reports the results and illustrate the stronger stability of the causality-aware approach (note the considerably smaller drop in performance between the no shift and shifted experiments), compared to the other methods that only adjust the training data. Further details about these experiments are presented in the Supplementary Section 12.
\vspace{-0.5cm}
\section{Final remarks}
\vskip -0.1in

Selection bias represents a common and challenging issue in ML applications. Here, we show that, in the particular context of anticausal tasks, deconfounding both the training and test set features leads to more stable predictions under dataset shifts of $P(\bfC, Y)$ generated by unstable selection biases. We show that the expected MSE of linear predictions generated by adjustment approaches that fail to deconfound the test set features will still be unstable w.r.t. shifts in the association between $\bfC_{ts}$ and $Y_{ts}$, even when the ML models are trained with unconfounded data and there are no shifts in $Var(Y_{ts})$. This is an important observation that (we feel) is not well appreciated in the ML community.

We illustrate this point using the causality-aware adjustment, which deconfounds the test set features (without having access to the test set labels), in addition to the training set features. The approach, however, requires access to test set confounder data, $\bfC_{ts}$, and, therefore, cannot be leveraged in applications where $\bfC_{ts}$ is unavailable. We anticipate, however, that it will be particularly useful in diagnostic health applications, where demographic risk factors such as gender, age, race, etc represent potential confounders and are likely recorded in the target (test) populations.

\section*{References}

\begin{enumerate}
\item Arjovsky M., Bottou L., Gulrajani I., Lopez-Paz D. (2019) Invariant risk minimization. \textit{arXiv:1907.02893v3}.

\item Arora, S., Venkataraman, V., Zhan, A., Donohue, S., Biglan, K. M, et al. Detecting and monitoring the symptoms of Parkinson's disease using smartphones: a pilot study. \textit{Parkinsonism \& Related Disorders}, \textbf{21}, 650-653 (2015).

\item Badgeley, M. A., Zech, J. R., Oakden-Rayner, L., Glicksberg, B. S., Liu, M., et al. Deep learning predicts hip fracture using confounding patient and healthcare variables. \textit{npj Digital Medicine}, 2:31, https://doi.org/10.1038/s41746-019-0105-1 (2019).

\item Bareinboim, E. and Pearl, J. (2012) Controlling selection bias in causal inference. AISTATS 2012.

\item Bickel, S., Bruckner, M., and Scheffer, T. (2009) Discriminative learning under covariate shift. \textit{Journal of Machine Learning Research}, \textbf{10}, 2137-2155.

\item Bot, B.M., et al. (2016) The mPower study, Parkinson disease mobile data collected using ResearchKit. \textit{Scientific Data} 3:160011 doi:10.1038/sdata.2016.11

\item Brestel, C., Shadmi, R., Tamir, I., Cohen-Sfaty, M., Elnekave, E. RadBot-CXR: Classification of four clinical finding categories in chest X-ray using deep learning. In \textit{MIDL 2018} (2018).

\item Calders T., Kamiran, F., Pechenizkiy, M. (2009) Building classifiers with independency constraints. ICDM Workshop on Domain Driven Data Mining.

\item Chaibub Neto, E., \textit{et al.} On the analysis of personalized medication response and classification of case vs control patients in mobile health studies: the mPower case study. arXiv:1706.09574 (2017).

\item Chaibub Neto, E., et al. (2019) Causality-based tests to detect the influence of confounders on mobile health diagnostic applications: a comparison with restricted permutations. In ML4H 2019 - Extended Abstract. arXiv:1911.05139.

\item Chaibub Neto, E. (2020a) Towards causality-aware predictions in static anticausal machine learning tasks: the linear structural causal model case. In Causal Discovery \& Causality-Inspired Machine Learning (CDML 2020) Workshop. \textit{arXiv:2001.03998} (accepted).

\item Chaibub Neto, E. (2020b) Causality-aware counterfactual confounding adjustment for feature representations learned by deep models. \textit{arXiv:2004.09466}


\item Chan, R., Jankovic, F., Marinsek, N., Foschini, L., Kourtis, L., et al. Developing measures of cognitive impairment in the real world from consumer-grade multimodal sensor streams. In \textit{KDD 2019} (2019).

\item Dai B, Ding S, and Wahba G. 2013. Multivariate Bernoulli distribution. \textit{Bernoulli} \textbf{19}: 1465-1483.

\item Dudik, M., Phillips, S. J., and Schapire, R. E. (2006) Correcting sample selection bias in maximum entropy density estimation. NeurIPS 2006.


\item Evers, L. J. W, Raykov, Y. P., Krijthe, J. H., de Lima, A. L. S., Badawy, R., et al  Real-life gait performance as a digital biomarker for motor fluctuations: the Parkinson@Home validation study. \textit{Journal of Medical Internet Research}, e19068 (2020).

\item Fong, C., Hazlett, C., and Imai, K. (2018).  Covariate balancing propensity score for a continuoustreatment: Application to the efficacy of political advertisements. The Annals of Applied Statistics,12(1), 156-177.

\item Gretton, A., Smola, A. J., Huang, J., Schmittfull, M.,Borgwardt, K. M., and Scholkopf, B. (2009). Covariate shift by kernel mean matching.  In Quinonero-Candela,  et al., editors, \textit{Dataset  Shift  in  Machine Learning}, 131-160. The MIT Press.

\item Heckman, J. J. (1979) Sample selection bias as a specification error. \textit{Econometrica}, \textbf{47}, 153-161.

\item Hirano, K. and Imbems, G. W. (2004). The propensity score with continuous treatments. In Applied Bayesian Modeling and Causal Inference from Incomplete-Data Perspectives: An Essential Journey with Donald Rubin's Statistical Family 73–84. Wiley, New York.

\item Huang, J., et al (2007) Correcting sample selection bias by unlabeled data. In NeurIPS 2007.

\item Kamiran, F. and Calders, T. (2012)  Data preprocessing techniques for classification without discrimination. \textit{Knowledge and Information Systems}, \textbf{33}, 1-33.

\item Kuang, K., Cui, C., Athey, S., Xiong, R., Li, B. (2018) Stable prediction across unknown environments. In \textit{SIGKDD 2018}.

\item Kuang, K., Xiong, R., Cui, C., Athey, S., Li, B. (2020) Stable prediction with model misspecification and agnostic distribution shift. \textit{arXiv:2001.11713}.

\item Magliacane, S., van Ommen, T., Claassen, T., Bongers, S., Versteeg, P., and Mooij, J. M. (2018). Domain adaptation by using causal inference to predict invariant conditional distributions. \textit{NeurIPS 2018}.

\item Pearl, J. (2009) \textit{Causality: models, reasoning, and inference.} Cambridge University Press New York, NY, 2nd edition.


\item Pearl, J., Glymour, M., Jewell, N. P. (2016) \textit{Causal inference in statistics: a primer.} Wiley.

\item Peters, J., Buhlmann, P., Meinshausen, N. (2016) Causal inference using invariant prediction: identification and confidence intervals. \textit{Journal of the Royal Statistical Society, series B}, \textbf{78}, 947-1012.

\item Quinonero-Candela, J., Sugiyama, M., Schwaighofer, A., and Lawrence, N. D. (2009). \textit{Dataset shift in machine learning.} MIT Press.

\item Liu, A. and Ziebart, B. (2014) Robust classification under sample selection bias. NeurIPS 2014.

\item Shimodaira H. (2000) Improving predictive inference under covariate shift by weighting the log-likelihood function. \textit{Journal of Statistical Planning and Inference}, \textbf{90}, 227-244.

\item Sieberts, S. K., et al (2020) Crowdsourcing digital health measures to predict Parkinson's disease severity: the Parkinson's Disease Digital Biomarker DREAM Challenge. https://doi.org/10.1101/2020.01.13.904722

\item Scholkopf B, Janzing D, Peters J, et al. (2012) On causal and anticausal learning. ICML 2012, 1255-1262.

\item Sugiyama, M., Krauledat, M., and MAzller, K. R. (2007).   Covariate  shift  adaptation  by  importance weighted cross-validation. \textit{Journal of Machine Learning Research}, \textbf{8}, 985-1005.

\item Subbaswamy A., Saria, S. (2018) Counterfactual normalization: proactively addressing dataset shift and improving reliability using causal mechanisms. \textit{UAI 2018}.

\item Subbaswamy A., Saria, S. (2020) From development to deployment: dataset shift, causality, and shift-stable models in health AI. \textit{Biostatistics}, \textbf{2}, 345-352.

\item Wang, M., Ge, W., Apthorp, D., Suominen, H. Robust feature engineering for parkinson disease diagnosis: new machine learning techniques. \textit{JMIR Biomedical Engineering}, \textbf{5}, e13611 (2020).

\item Wright, S. (1934) The method of path coefficients. \textit{The Annals of Mathematical Statistics}, \textbf{5}:161-215.
\end{enumerate}

\section*{Author contributions}

Conceptualization, methodology, and analysis: ECN. Data curation and resources: PS, SKS, and LO. Writing: ECN. Review and editing: PS, SKS, and LO.

\clearpage

\beginsupplement

\noindent {\huge SUPPLEMENT}

\section{Hypothetical illustrative example of selection bias}

Consider an hypothetical scenario where we are interested in developing a ML model for predicting disease severity (represented by the outcome variable, $Y$) using some of the disease symptoms (represented by $\bfX$) as the input variables. Suppose that a given demographic variable (such as gender, age, or race) is a risk factor for the disease, and also affects the disease symptoms. In this scenario, the demographic variable (represented by $C$) will be a confounder of the disease/symptoms relationship.

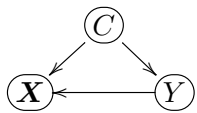
\begin{figure}[!h]
$$
\xymatrix@-1pc{
& *+[F-:<10pt>]{C} \ar[dl] \ar[dr] & \\
*+[F-:<10pt>]{\bfX}  & & *+[F-:<10pt>]{Y} \ar[ll] \\}
$$
\caption{Biological aspects of the data generation process.}
\label{fig:biological.causal.graph}
\end{figure}

For concreteness, suppose the confounder is race and the disease severity status of each patient can be classified as ``mild" or ``severe". A causal graph representation of this data generation process is given in Figure \ref{fig:biological.causal.graph}, where the arrow $C \rightarrow Y$ indicates that race is a risk factor for the disease, the arrow $Y \rightarrow \bfX$ indicates that the disease causes the symptoms, and the arrow $C \rightarrow \bfX$ indicates that race also directly affects the symptoms.

Note that the above graph represents the causal relations that are governed by biological aspects of the data generation process. For instance, the arrow $C \rightarrow Y$ indicates that people from different races have different risks to develop the disease. In this work, we assume that these biological aspects are stable across different training and test sets (e.g., we assume that the risk of white individuals in the training set is the same as the risk of white individuals in any distinct test set).

\begin{figure}[!h]
$$
\xymatrix@-1pc{
(a) & *+[F-:<10pt>]{C} \ar[dl] \ar[dr] \ar[r] & *+[F]{S_A} & (b) & *+[F-:<10pt>]{C} \ar[dl] \ar[dr] \ar[r] & *+[F]{S_B} \\
*+[F-:<10pt>]{\bfX}  & & *+[F-:<10pt>]{Y} \ar[ll] \ar[u] & *+[F-:<10pt>]{\bfX}  & & *+[F-:<10pt>]{Y} \ar[ll] \ar[u] \\}
$$
\caption{Data generation processes affected by selection biases.}
\label{fig:selection.bias.causal.graph}
\end{figure}
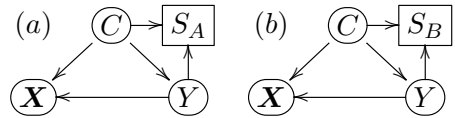

Now, suppose that the training data comes from hospital $A$ located in a city whose population is predominantly white. Suppose, as well, that hospital $A$ is the best equipped hospital in the city to treat severe cases of the disease. (So, that most severe cases tend to be sent to this hospital). In this scenario we might see an over-representation of white patients with severe disease in the training set. This additional source of association between race and disease status is represented by the auxiliar variable $S_A$ in the graph in Figure \ref{fig:selection.bias.causal.graph}a, which represents the data generation process biased by the selection mechanism operating at hospital $A$.
Here, $S_A$ represents a binary variable assuming the value 1 when the data is generated in hospital $A$, and 0 otherwise. The squared frame around $S_A$ indicates that the data generation process giving rise to the data is conditional on the fact that $S_A = 1$, that is, on the fact that the data is comming from hospital $A$. Note that, application of the d-separation criterion~\cite{pearl2009} to the causal graph in Figure \ref{fig:selection.bias.causal.graph}a shows that, because $S_A$ is a collider, we have that conditional on $S_A = 1$ the additional path $C \rightarrow S_A \leftarrow Y$ is open and, therefore, contributes to the association between $C$ and $Y$. This additional association, however, is usually different across different cities, communities, and hospitals. For instance, consider a hospital $B$ located in a different city whose population is composed predominantly of non-white individuals. Suppose, as well, that hospital $B$ is also the best equipped hospital in town to handle the more severe cases. Now, for hospital $B$ we have a different selection mechanism at work (Figure \ref{fig:selection.bias.causal.graph}b), and we would expect to see an over-representation of non-white patients with severe cases of the disease, so that the association between race and disease severity is flipped in this second subpopulation.

In other words, the associations generated by selection mechanisms tend to be unstable across different hospitals/subpopulations. As a consequence, a classifier trained on the data generated by hospital $A$ might fail to generalize when deployed in a different subpopulation where the association between $C$ and $Y$ is flipped relative to the training set.

\section{Simple balancing adjustments when the test set population is known a priori}

\begin{figure}[!h]
\centerline{\includegraphics[width=2.8in]{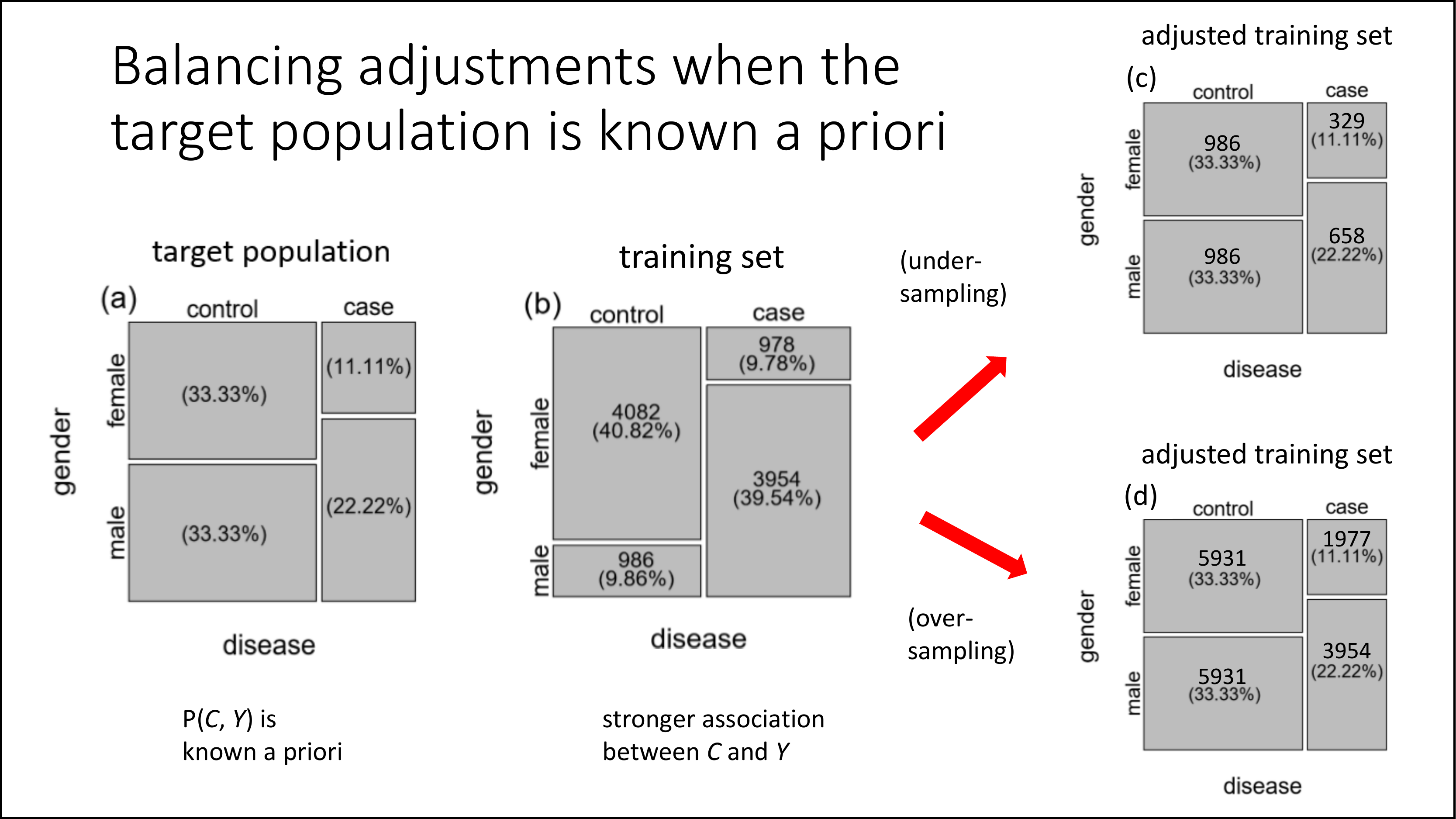}}
\vskip -0.1in
\caption{Balancing adjustments.}
\label{fig:balancing.adjustments}
\end{figure}

In situations where the target (test set) population where the ML model will be deployed is known a priori, simple balancing adjustments can be used to make sure that the joint distribution $P(\bfC, Y)$ is the same in the training and test sets. As an illustration, consider a confounded classification task where the goal is to classify disease status (case vs control) and gender is a risk factor for the disease. Suppose that it is known, a priori, that a disease affects one third of the population and is two times more common in males than in females in the target population. The mosaic plot in Figure \ref{fig:balancing.adjustments}a describes the joint distribution of gender and disease status in the target population. Suppose, as well, that we have access to a training set containing 10,000 samples, but that, due to selection mechanisms, gender and disease status are more strongly associated in the training dataset than in the target population. Figure \ref{fig:balancing.adjustments}b shows a mosaic plot describing the joint distribution of gender and disease status in the development dataset. In this situation we can correct the shift in the training set distribution $P(C_{tr}, Y_{tr})$ relative to the test set $P(C_{ts}, Y_{ts})$ by simply rebalancing the training data to match the test set distribution. We can for instance apply a matching procedure where we randomly undersample from the majority classes in training set in order to obtain an adjusted dataset that matches the label/counfounder class proportions in the target population (\ref{fig:balancing.adjustments}c). Similarly, we can over-sample from the minority classes in order to match the class proportions in the target population (\ref{fig:balancing.adjustments}d). (Or still apply a combination of under- and over-sampling to match the proportions in the target population.)

\section{Related work}

As clearly articulated by Subbaswamy and Saria (2020) there are, broadly speaking, two types of stable prediction approaches: (i) \textit{reactive} methods, that use data (or knowledge) from the intended deployment/target population to correct for shifts; and (ii) \textit{proactive} methods, that do not require data from the deployment/target populations, and are able to learn models that are stable with respect to unknown dataset shifts.

Many reactive approaches in the literature (Shimodaira 2000; Sugiyama et al 2007; Dudik, Phillips, and Schapire 2006; Huang et al 2007; Gretton et al 2009; Bickel, Bruckner, and Scheffer 2009; Liu and Ziebart 2014) deal with dataset shift by reweighting the training data to make it more closely aligned with the target test distribution. In this paper, however, we focus on anticausal prediction tasks (Scholkopf et al 2012) and address only dataset shifts in the joint distribution of the confounders and outcome variable, $P(\bfC, Y)$, caused by selection biases (Heckman 1979; Bareinboim and Pearl 2012). In our particular context, we can still use simple reactive approaches when the target (test set) joint distribution, $P(\bfC_{ts}, Y_{ts})$ is known. For instance, if we know, a priori, the prevalence of a disease with respect to a given demographic risk factor in the target population, then we can either subsample or oversample the training data in order to make the training set distribution $P(\bfC_{tr}, Y_{tr})$ as close as possible $P(\bfC_{ts}, Y_{ts})$. (Figure \ref{fig:balancing.adjustments} provides an illustrative example.) In classification tasks, simple balancing approaches, such as matching or approximate inverse probability weighting, can be used to subsample or oversample the training data. In regression tasks, approaches such as propensity scores for continuous variables (Hirano and Imbems 2004), covariate balancing propensity score methods for continuous variables (Fong, Hazlett, and Imai 2018), or standard propensity score matching applied to dichotomized outcome data, can be used.

The more challenging case where we face unknown shifts in $P(\bfC_{ts}, Y_{ts})$ (the case we address in this paper) requires more sophisticated adjustment approaches. Several proactive methods have been proposed in the literature. For instance, invariant learning approaches (Peters et al 2016; Rojas-Carulla et al 2018; Magliacane et al 2018; Arjovsky et al 2019) employ multiple training datasets in order to learn invariant predictions. The causality-aware approach (adopted in this paper), on the other hand, only requires a single training set.

Another proactive approach, which can be applied to anticausal tasks based on a single training set, is the counterfactual normalization method proposed by Subbaswamy and Saria (2018). The approach requires full knowledge of the causal graph describing the data generation process and is implemented in several steps. First, it identifies a set vulnerable variables that make the ML model susceptible to learning unstable relationships that might lead to poor generalization across shifted dataset. Second, the approach performs a node-splitting operation in order to augment the causal graph with counterfactual variables which isolate unstable paths of statistical associations and allow the retention of some stable paths involving vulnerable variables. Third, the approach determines a stable set of input variables that can be used to train a more stable ML model. In practice, the approach is implemented with linear (or additive) models.

Similarly to counterfactual normalization, the causality-aware approach also leverages counterfactual features to improve stability and is also implemented with linear (or additive) models. There are, nonetheless, important differences. The key idea (in the context of anticausal prediction tasks) is to train and evaluate supervised ML algorithms on counterfactually simulated data which retains only the associations generated by the causal influences of the output variable on the inputs. Noteworthy, as described in detail in Chaibub Neto (2020a), in the particular context of linear structural causal models, it is always possible to reparameterize the model in a way that the covariance among the features and among the confounders is pushed towards the respective error terms. This allows the generation of counterfactual features without even knowing the causal relations among features and the causal relations among the confounders. As a consequence, the causality-aware approach does not require knowledge of the full data generation process (at least for linear models). Contrary to counterfactual normalization, where the full causal diagram needs to be specified, the causality-aware approach only requires knowledge of which variables are confounders.

Finally, the methods proposed by Kuang et al (2018) and Kuang et al (2020) represent another set of stable prediction approaches. The key idea behind these methods is to find a set of covariates for which the expected value of the outcome is stable across distinct test set environments. These covariates fall into two classes: stable variables ($\bfS$) that have an structural relationship with the outcome, and unstable variables ($\bfV$) that can be associated with both the outcome and the stable variables but do not have a causal relation with the outcome. Assuming that there exists a stable function $f(\bfs)$ such that for all testing environments $E(Y \mid \bfS = \bfs, \bfV = \bfv) = E(Y \mid \bfS = \bfs) = f(\bfs)$ - a condition which is fulfilled when $Y \ci \bfV \mid \bfS$ - the approach is able to learn the stable function $f(\bfs)$ without prior knowledge about which variables are stable or unstable. These methods, however, are tailored to causal prediction tasks (i.e., where the inputs have a causal effect on the outcome), and cannot be directly applied in anticausal tasks\footnote{Note that in anticausal prediction tasks $\bfS$ might be a collider. Hence, if $\bfS$ is a collider, it follows that conditional on $\bfS$, $Y$ cannot be independent of $\bfV$, and the assumption $Y \ci \bfV \mid \bfS$ cannot hold}.

\subsection{A note on the use of data balancing approaches to combat selection biases}

Matching and data reweighting approaches are widely used confounding adjustment methods for dealing with selection biases in health related ML applications (likely due to their non-parametric nature and easy of implementation). For instance, in mobile health studies, the development data is often split into i.i.d. training and test sets, and then matching (or any other sort of data reweighting technique) is applied to both the training and test sets (Arora et al. 2015, Chaibub Neto et al. 2017, Brestel et al. 2018, Chan et al. 2019, Badgley et al. 2019, Evers et al. 2020, Wang et al. 2020, Sieberts et al. 2020).

While training a classifier on matched data (where the joint distribution of the confounders and the labels is perfectly balanced) can prevent it from learning the confounding signal, this does not imply that the predictions will be stable. As described in Section 3 in the main text (and in Section 10 in the Supplement), because ML predictions are a combination of both the trained model and the test set features we have that classifiers trained with matched data will still show unstable predictive performance when deployed in shifted target populations since the input data still retains the confounding signal. While the application of data balancing approaches to both the training and test data can be used to estimate the predictive performance that can be achieved by a stable ML algorithm, it is important to clarify that this practice does not actually produce stable learners, since matching and data re-weighting do not deconfound the learner inputs. As illustrated by our experiments in Sections 4 and 5 of the main text (and Sections 11 and 12 in the Supplement), the deployment of classifiers trained with balanced data will still be unstable under dataset shifts.

\section{Expected MSE for arbitrary anticausal prediction tasks based on linear models}

\begin{figure}[!h]
$$
\xymatrix@-2.2pc{
& U_{C_1} \ar[drr] \ar@/^1pc/@{<->}[rr] & \ldots & U_{C_m} \ar[drr] &&& \\
U_{X_1} \ar[ddr] \ar@/_1pc/@{<->}[dd] & & & *+[F-:<10pt>]{C_1} \ar[ddll] \ar[ddddll] \ar[dddrrr] & \ldots & *+[F-:<10pt>]{C_m} \ar[ddllll] \ar[ddddllll] \ar[dddr] \\
\vdots &&&&&&  \\
U_{X_p} \ar[ddr] & *+[F-:<10pt>]{X_1} &&&& \\
& \vdots & & & & & *+[F-:<10pt>]{Y} \ar[ulllll] \ar[dlllll] \\
& *+[F-:<10pt>]{X_p} &&& U_Y \ar[urr] && \\
}
$$
\vskip -0.1in
  \caption{Confounded anticausal prediction task example.}
  \label{fig:confounded.anticausal.example}
  \vskip -0.1in
\end{figure}
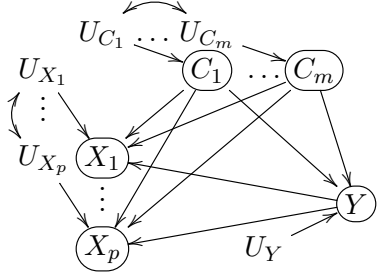

Consider the arbitrary anticausal prediction task model in Figure \ref{fig:confounded.anticausal.example},
where the double arrows connecting the variables $\{U_{X_1}, \ldots, U_{X_p}\}$ (and $\{U_{C_1}, \ldots, U_{C_m}\}$) represent the fact that these error terms are correlated\footnote{Note that the above model might represent a reparameterization of a model with uncorrelated error terms and unknown causal relations among the $\bfX$ input variables, as well as, among the $\bfC$ confounder variables. As described in detail in Chaibub Neto (2020a), for linear structural equation models, we can always reparameterize the original model in a way where the covariance structure among the input variables, as well as, the covariance structure among the confounder variables is pushed towards the respective error terms as illustrated in Figure \ref{fig:confounded.anticausal.example}.}. Without loss of generality assume that the data has been centered, so that the linear structural causal models describing the data generation process are given by,
\begin{align}
C_j &= U_{C_j}~, \\
Y &= \sum_i \beta_{Y{C_i}} \, C_i + U_Y~, \\
X_j &= \beta_{{X_j}Y} Y + \sum_i \beta_{{X_j}{C_i}} \, C_i + U_{X_j}~, \label{eq:feature.model}
\end{align}
for $j = 1, \ldots, p$ and $i = 1, \ldots, m$. The causality-aware features are estimated as,
\begin{align}
\hat{X}_j^\ast &= X_j - \sum_i \hat{\beta}_{{X_j}{C_i}} C_i~,
\end{align}
and converge to,
\begin{align}
X_j^\ast &= X_j - \sum_i \beta_{{X_j}{C_i}} C_i \\
&= \beta_{{X_j}Y} Y + U_{X_j}~, \label{eq:feature.model.ca}
\end{align}
asymptotically.

Now, let $\hat{Y} = \bfX_{ts} \hat{\bfBeta}^{tr}$ represent the prediction of a linear regression model, where $\bfX_{ts}$ represents the test set features, and $\hat{\bfBeta}^{tr}$ represents the regression coefficients estimated from the training set. By definition, the expected mean squared error of the prediction is given by,
\begin{align*}
E&[MSE] = E[(Y_{ts} - \hat{Y})^2] = \\
&= E[Y^2_{ts}] + E[\hat{Y}^2] - 2 E[\hat{Y} Y_{ts}] \\
&= Var(Y_{ts}) + E[\hat{Y}^2] - 2 Cov(\hat{Y}, Y_{ts})~,
\end{align*}
since $E[Y_{ts}] = 0$. Direct computation shows that,
\begin{align*}
E[\hat{Y}^2] &= E[(\sum_{j=1}^{p} X_{j,ts} \hat{\beta}_j^{tr})^2] \\
&= \sum_{j=1}^{p} (\hat{\beta}_j^{tr})^2 Var(X_{j,ts}) \, + \\
&\;\;\;\;+ 2 \sum_{j < k} \hat{\beta}_j^{tr} \hat{\beta}_k^{tr} Cov(X_{j,ts}, X_{k,ts})~,
\end{align*}
and,
\begin{align*}
Cov(\hat{Y}, Y_{ts}) &= \sum_{j=1}^{p} \hat{\beta}_j^{tr} Cov(X_{j,ts}, Y_{ts})~,
\end{align*}
so that,
\begin{align*}
E&[MSE] \, =  \, Var(Y_{ts}) \, + \\
&\;\;\;\;+ \sum_{j=1}^{p} (\hat{\beta}_j^{tr})^2 Var(X_{j,ts}) \, + \\
&\;\;\;\;+ 2 \sum_{j < k} \hat{\beta}_j^{tr} \hat{\beta}_k^{tr} Cov(X_{j,ts}, X_{k,ts}) \, - \\
&\;\;\;\;- 2 \sum_{j=1}^{p} \hat{\beta}_j^{tr} Cov(X_{j,ts}, Y_{ts})~.
\end{align*}

Next, we derive the expressions for $Var(X_{j,ts})$, $Cov(X_{j,ts}, X_{k,ts})$, and $Cov(X_{j,ts}, Y_{ts})$ and show that they still depend on $Cov(Y_{ts}, C_{i,ts})$. From equation (\ref{eq:feature.model}) we have that,
\begin{align*}
&Var(X_{j,ts}) = \\
&= Var(\beta_{{X_j}Y} \, Y_{ts} + \sum_i \beta_{{X_j}{C_i}} \, C_{i,ts} + U_{X_j}) \\
&= \sigma^2_{X_j} + \beta_{{X_j}Y}^2 \, Var(Y_{ts}) + \\
&\;\;\;\; + \sum_i \beta_{{X_j}{C_i}}^2 \, Var(C_{i,ts}) \, + \\
&\;\;\;\; + 2 \, \sum_{i < i'} \beta_{{X_j}{C_i}} \, \beta_{{X_j}{C_{i'}}} \, Cov(C_{i,ts}, C_{i',ts}) \, + \\
&\;\;\;\; + \, 2 \, \beta_{{X_j}Y} \sum_i \beta_{{X_j}{C_i}} \, Cov(Y_{ts}, C_{i,ts})~, \\
\end{align*}
\vskip -0.5in
\begin{align*}
&Cov(X_{j,ts}, X_{k,ts}) = \\
&= Cov(\beta_{{X_j}Y} \, Y_{ts} + \sum_i \beta_{{X_j}{C_i}} \, C_{i,ts} + U_{X_j}, \\
&\hspace{1.3cm} \beta_{{X_k}Y} \, Y_{ts} + \sum_i \beta_{{X_k}{C_i}} \, C_{i,ts} + U_{X_k}) \\
&= \beta_{{X_j}Y} \, \beta_{{X_k}Y} \, Var(Y_{ts}) + \\
&\;\;\;\;+ \beta_{{X_j}Y} \, \sum_i \beta_{{X_k}{C_i}} \, Cov(Y_{ts}, C_{i,ts}) + \\
&\;\;\;\;+ \beta_{{X_k}Y} \, \sum_i \beta_{{X_j}{C_i}} \, Cov(Y_{ts}, C_{i,ts}) + \\
&\;\;\;\;+ \sum_i \sum_{i'} \beta_{{X_j}{C_i}} \, \beta_{{X_k}{C_{i'}}} \, Cov(C_{i,ts}, C_{i',ts}) + \\
&\;\;\;\;+ Cov(U_{X_j}, U_{X_k})
\end{align*}
\begin{align*}
&Cov(X_{j,ts}, Y_{ts}) = \\
&= Cov(\beta_{{X_j}Y} \, Y_{ts} + \sum_i \beta_{{X_j}{C_i}} \, C_{i,ts} + U_{X_j}, Y_{ts}) \\
&= \beta_{{X_j}Y} \, Var(Y_{ts}) + \sum_i \beta_{{X_j}{C_i}} \, Cov(Y_{ts}, C_{i,ts})~,
\end{align*}
showing that these three quantities still depend on $Cov(Y_{ts}, C_{i,ts})$ (in addition to depending on $Var(Y_{ts})$, $Var(C_{ts})$, and $Cov(C_{i,ts}, C_{i',ts})$). This observation implies that the $E[MSE]$ will still be unstable w.r.t. shifts in these quantities, even when the regression model is trained in unconfounded data (a situation where the estimates $\hat{\beta}_j^{tr}$ are not influenced by spurious associations generated by the confounders). This explains why it is not enough to deconfound the training features alone. While training a regression model using deconfounded features allows us to estimate deconfounded model weights,  $\hat{\bfBeta}^{tr}$, the prediction $\hat{Y} = \bfX_{ts} \hat{\bfBeta}^{tr}$ is a function of both the trained model $\hat{\bfBeta}^{tr}$ and the test set features, $\bfX_{ts}$. As a consequence, if we do not deconfound the test set features, the expected MSE will still be influenced by the confounders (since, in anticausal prediction tasks, the features are functions of both the confounder and outcome variables).

The expected MSE of models trained with test set features processed according to the causality-aware approach, on the other hand, do not depend on $Cov(Y_{ts}, C_{i,ts})$, $Var(C_{ts})$, or $Cov(C_{i,ts}, C_{i',ts})$, since the approach also deconfounds the test set features. Note that direct computation of $Var(X_{j,ts}^\ast)$, $Cov(X_{j,ts}^\ast, X_{k,ts}^\ast)$, and $Cov(X_{j,ts}^\ast, Y_{ts})$ based on the causality-aware features, $X_{j,ts}^\ast = \beta_{{X_j}Y} \, Y_{ts} + U_{X_j}$, shows that,
\begin{align*}
Var(X_{j,ts}^\ast) &= Var(\beta_{{X_j}Y} \, Y_{ts} + U_{X_j}) \\
&= \sigma^2_{X_j} + \beta_{{X_j}Y}^2 \, Var(Y_{ts})~,
\end{align*}
\begin{align*}
&Cov(X_{j,ts}^\ast, X_{k,ts}^\ast) = \\
&= Cov(\beta_{{X_j}Y} \, Y_{ts} + U_{X_j}, \beta_{{X_k}Y} \, Y_{ts} + U_{X_k}) \\
&= \beta_{{X_j}Y} \, \beta_{{X_k}Y} \, Var(Y_{ts}) + Cov(U_{X_j}, U_{X_k})~,
\end{align*}
\begin{align*}
Cov(X_{j,ts}^\ast, Y_{ts}) &= Cov(\beta_{{X_j}Y} \, Y_{ts} + U_{X_j}, Y_{ts}) \\
&= \beta_{{X_j}Y} \, Var(Y_{ts})~,
\end{align*}
no longer depend on $Cov(Y_{ts}, C_{i,ts})$, $Var(C_{ts})$, or $Cov(C_{i,ts}, C_{i',ts})$, so that the approach will be stable against shifts in these quantities. Observe, nonetheless, that it will still be influenced by shifts in $Var(Y_{ts})$. (We point out, however, that the dependence of $E[MSE]$ on $Var(Y_{ts})$ is, in general, unavoidable since, by definition, $E[MSE]= Var(Y_{ts}) + E[\hat{Y}^2] - 2 Cov(\hat{Y}, Y_{ts})$.)

\section{Additional details - synthetic data experiments}

Here, we present synthetic data experiments for both regression and classification tasks. In both cases we generate data from an anticausal prediction task involving five features and one confounder variable. We describe next, the simulation details for each of these tasks.

\subsection{Regression task illustrations}

For the regression task illustrations we simulate data from the model,
$$
\xymatrix@-1.2pc{
& & & *+[F-:<10pt>]{C} \ar[dll]_{\beta_{{X_1}C}} \ar[ddll] \ar[ddddll] \ar[ddr]^{\beta_{YC}} & U_{C} \ar[l] \ar@/^1pc/@{<->}[ddr] & \\
U_{X_1} \ar[r] \ar@/_1pc/@{<->}[dd] \ar@/_1.5pc/@{<->}[ddd] & *+[F-:<10pt>]{X_{1}} &&&& \\
& *+[F-:<10pt>]{X_{2}} & & & *+[F-:<10pt>]{Y} \ar[ulll] \ar[lll] \ar[ddlll]^{\beta_{{X_{5}}Y}} & U_Y \ar[l] \\
U_{X_{2}} \ar[ru] & \vdots &&&& \\
U_{X_{5}} \ar[r] & *+[F-:<10pt>]{X_{5}} &&&  \\
}
$$
where we change the covariance of the error terms $U_C$ and $U_Y$ in order to simulate the effects of selection biases in the joint distribution $P(C, Y)$.

The model is described by the following set of linear structural causal equations,
\begin{align}
C &= U_{C}~, \label{eq:C.regr.model} \\
Y &= \beta_{YC} \, C + U_Y~, \label{eq:Y.regr.model} \\
X_{j} &= \beta_{{X_j}Y} \, Y + \beta_{{X_j}{C}} \, C + U_{X_j}~, \label{eq:X.regr.model}
\end{align}
for $j = 1, \ldots, 5$, and where the error terms $U_C$ and $U_Y$ are distributed according to,
\begin{equation}
\begin{pmatrix}
U_{C} \\
U_{Y} \\
\end{pmatrix} \,
\sim \, \mbox{N}_2\left(
\begin{pmatrix}
0 \\
0 \\
\end{pmatrix}\, , \,
\begin{pmatrix}
\phi_{CC} & \phi_{CY} \\
\phi_{CY} & \phi_{YY} \\
\end{pmatrix} \right)~, \label{eq:errors.C.Y.distr}
\end{equation}
and $\bfU_X = (U_{X_1}, \ldots, U_{X_{5}})^T$ is distributed according to a multivariate normal distribution,
\begin{equation}
\bfU_X \, \sim \, N_{5}(\bfzero \, , \ \bfSigma_{\bfU_X})~, \label{eq:errors.X.distr}
\end{equation}
where the $ij$th entry of the covariance matrix $\bfSigma_{\bfU_X}$ is given by 1 for $i = j$, and by $\rho^{|i-j|}$ for $i \not= j$.

Note that, for the above model, we have that,
\begin{align}
Var&(C) = \phi_{CC}~, \\
Cov&(Y, C) = Cov(\beta_{YC} \, C + U_Y, C) \nonumber \\
&= \beta_{YC} \, Var(C) + Cov(U_Y, C) \nonumber \\
&= \beta_{YC} \, \phi_{CC} + \phi_{CY}~, \\
Var&(Y) = Var(\beta_{YC} \, C + U_Y) \nonumber \\
&= \beta_{YC}^2 \, Var(C) + Var(U_Y) + \nonumber \\
&\;\;\;\;+ 2 \, \beta_{YC} \, Cov(C, U_Y) \nonumber \\
&= \beta_{YC}^2 \, \phi_{CC} + \phi_{YY} + 2 \, \beta_{YC} \, \phi_{CY}~,
\end{align}
so that for fixed values of $Var(C)$, $Cov(Y, C)$, $Var(Y)$, and $\beta_{{Y}{C}}$ we can determine the values of $\phi_{CC}$, $\phi_{CY}$, and $\phi_{YY}$ as follows,
\begin{align}
\phi_{CC} &= Var(C)~, \label{eq:phi.CC} \\
\phi_{CY} &= Cov(Y, C) - \beta_{YC} \, Var(C)~, \label{eq:phi.CY} \\
\phi_{YY} &= Var(Y) - \beta_{YC}^2 \, Var(C) - \nonumber \\
&\;\;\;\;-2 \, \beta_{YC} \, Cov(Y, C)~. \label{eq:phi.YY}
\end{align}


In our experiments, we simulate data as follows:
\begin{enumerate}
\item Sample the simulation parameters $\beta_{{X_j}Y}$, $\beta_{{X_j}{C}}$, and $\beta_{{Y}{C}}$ from a $U(-1, 1)$ distribution, and $\rho$ from a $U(-0.5, 0.5)$ distribution.
\item Given the fixed values for $Var(C)$, $Cov(Y, C)$, and $Var(Y)$, and the sampled value for $\beta_{{Y}{C}}$, we compute $\phi_{CC}$, $\phi_{CY}$, and $\phi_{YY}$ as described in equations (\ref{eq:phi.CC}), (\ref{eq:phi.CY}), and (\ref{eq:phi.YY}).
\item Sample the error terms $U_C$ and $U_Y$ according to (\ref{eq:errors.C.Y.distr}), and then simulate the $C$ and $Y$ data according to equations (\ref{eq:C.regr.model}) and (\ref{eq:Y.regr.model}), respectively.
\item Sample the error terms $\bfU_X$ according to (\ref{eq:errors.X.distr}), and then simulate the feature data according to equation (\ref{eq:X.regr.model}).
\end{enumerate}

In order to illustrate the influence of $Var(Y_{ts})$ in the stability of the predictions, we performed two regression task experiments. In the first, (whose results were presented in the main text) we kept the $Var(Y_{ts})$ constant across the test sets. In the second, we increased $Var(Y_{ts})$ across the test sets. Each of these experiments were based on 1000 simulations. For each simulation replication we:
\begin{enumerate}
\item Generate the training set ($n = 1,000$) by setting $Var(C_{tr}) = 1$, $Cov(Y_{tr}, C_{tr}) = 0.8$, and $Var(Y_{tr}) = 1$ and then simulating the data as described above.
\item Generate 9 distinct test sets (each containing $n = 1,000$ samples). Each test set was generated with an increasing amount of shift in the $P(C, Y)$ distribution. In the first experiment this was accomplished by varying $Cov(Y_{ts}, C_{ts})$ according to $\{0.8$, 0.6, 0.4, 0.2, 0.0, -0.2, -0.4, -0.6, $-0.8\}$ across the 9 test sets, and by varying $Var(C_{ts})$ according to $\{1.00$, 1.25, 1.50, 1.75, 2.00, 2.25, 2.50, 2.75, $3.00\}$, while keeping $Var(Y_{ts})$ fixed at 1. In the second experiment, we varied $Cov(Y_{ts}, C_{ts})$ as before, but kept $Var(C_{ts})$ fixed at 1, while varying $Var(Y_{ts})$ according to $\{1.00$, 1.25, 1.50, 1.75, 2.00, 2.25, 2.50, 2.75, $3.00\}$ across the test sets.
\item For the causality-aware approach we: adjust the training set, and each of the 9 distinct test sets; fit a regression model on the adjusted training set data; use the same trained model to predict on the 9 adjusted test sets; and evaluate the test set performances using MSE.
\item For the ``poor man's" version of the causality-aware approach we: only adjust the training data; fit a regression model to the deconfounded training data; use the trained model to predict on the 9 unadjusted test sets; and evaluate the test set performances using MSE.
\item For the ``no adjustment" approach we: fit a regression model to the confounded training data; use the trained model to predict on the 9 unadjusted test sets; and evaluate the test set performances using MSE.
\end{enumerate}

Note that in our experiments, only $Cov(Y, C)$, $Var(C)$, and $Var(Y)$ may vary across the training and test sets. We still adopt the same values of the $\beta_{{X_j}Y}$, $\beta_{{X_j}{C}}$, $\beta_{{Y}{C}}$, and $\rho$ parameters in the simulation of the training and test sets, so that $P(\bfX \mid C, Y)$ is still stable across the datasets. Observe, as well, that the first test set is generated using the same values of $Cov(Y, C)$, $Var(C)$, and $Var(Y)$ as the training set, so that it illustrates the case where the training and test sets are i.i.d.

\begin{figure*}[!h]
\centerline{\includegraphics[width=\linewidth]{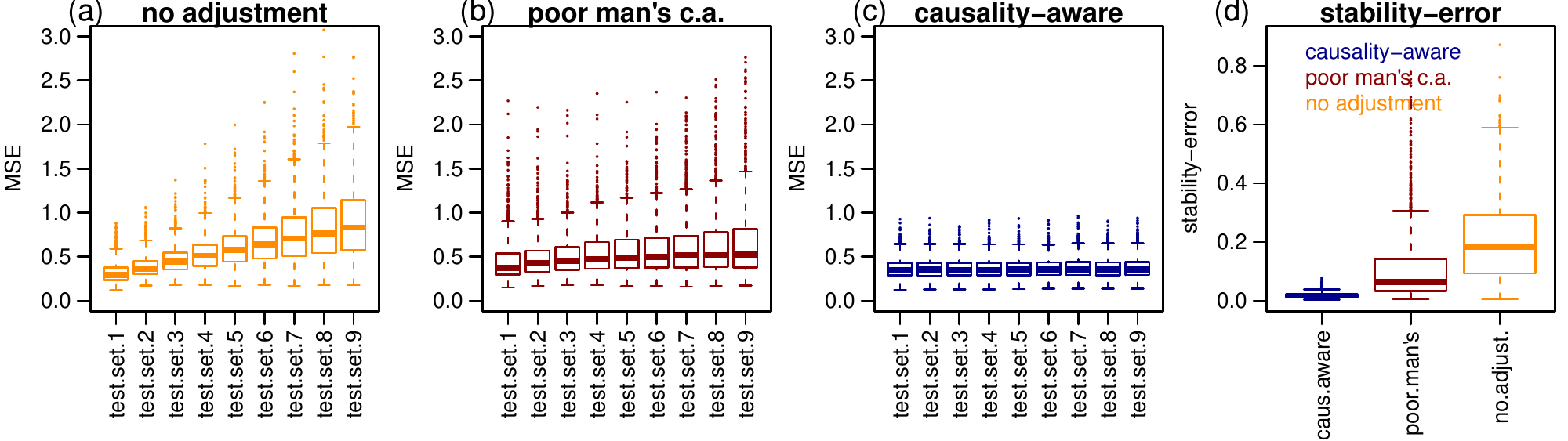}}
\vskip -0.15in
\caption{Regression task synthetic data experiments. Fixed $Var(Y_{ts})$ case.}
\label{fig:mpower.aucs.stability.regr.e1}
\end{figure*}

\begin{figure*}[!h]
\centerline{\includegraphics[width=\linewidth]{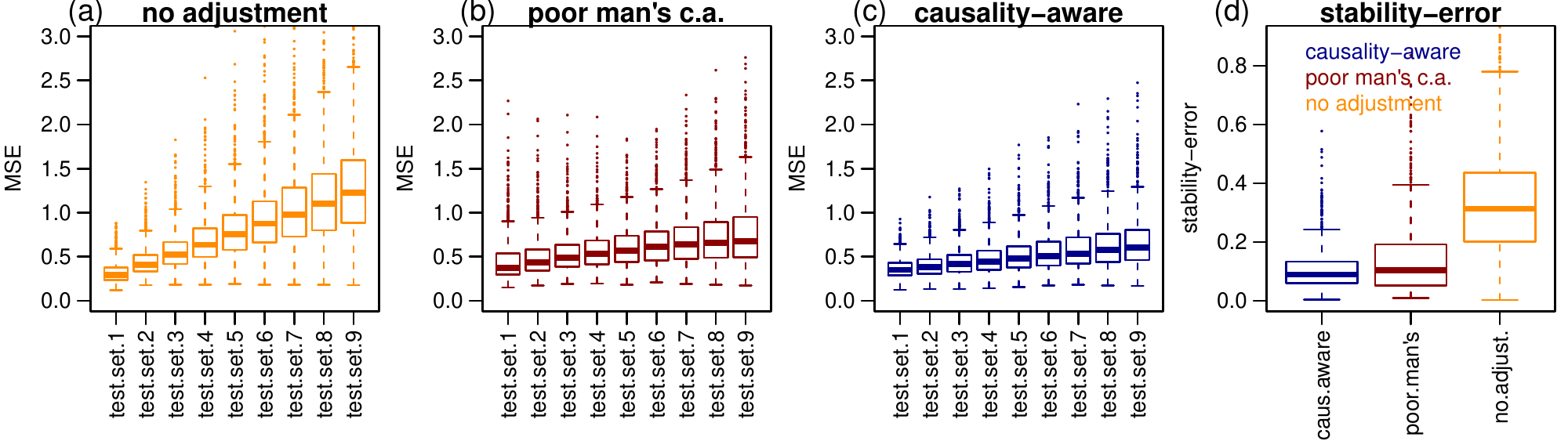}}
\vskip -0.15in
\caption{Regression task synthetic data experiments. Varying $Var(Y_{ts})$ case.}
\label{fig:mpower.aucs.stability.regr.e2}
\end{figure*}

Figure \ref{fig:mpower.aucs.stability.regr.e1} reports the results for the first experiment (these same results are also presented in Figure \ref{fig:synthetic.data.stability}a in the main text). Note that because we kept $Var(Y_{ts})$ constant across the test sets we see perfect stability for the causality-aware approach. (Note that varying $Cov(Y_{ts}, C_{ts})$ and $Var(C_{ts})$ has no influence on the stability of the results, since the expected MSE for the causality-aware approach only depends on $Var(Y_{ts})$.)

Figure \ref{fig:mpower.aucs.stability.regr.e2} reports results for the second experiment based on varying $Var(Y_{ts})$ values. As expected, we now observe instability in the causality-aware approach too. The causality-aware predictions, however, are still more stable than the predictions from the other approaches.

Panel d on Figures \ref{fig:mpower.aucs.stability.regr.e1} and \ref{fig:mpower.aucs.stability.regr.e2} present a comparison of the stability-errors, defined as the standard deviation of the MSE scores across the 9 test sets in each simulation replication. It again illustrates the better stability of the causality-aware approach.

Note that, for the particular causal graph used in our experiments, application of counterfactual normalization (Subbaswamy and Saria 2018) approach to deconfounded the training data would augment the causal graph with the counterfactual variables $X_{j}(C = \emptyset)$ (which correspond to the values of $X_j$ we would have seen, had $C$ not being a parent of $X_j$), and would return the counterfactual variables $X_{j}(C = \emptyset)$ as the stable set for predicting $Y$.
$$
\xymatrix@-1.4pc{
& & & *+[F-:<10pt>]{C} \ar@/_1pc/[dll] \ar@/_1.75pc/[ddll] \ar@/^2.25pc/[ddddll] \ar[ddr] & U_{C} \ar[l] \ar@/^1pc/@{<->}[ddr] & \\
U_{X_1} \ar[r] \ar@/_1pc/@{<->}[dd] \ar@/_1.5pc/@{<->}[ddd] & *+[F-:<10pt>]{X_{1}} & *+[F-:<10pt>]{X_{1}(C = \emptyset)} \ar[l] &&& \\
& *+[F-:<10pt>]{X_{2}} & *+[F-:<10pt>]{X_{2}(C = \emptyset)} \ar[l] & & *+[F-:<10pt>]{Y} \ar[ull] \ar[ll] \ar[ddll] & U_Y \ar[l] \\
U_{X_{2}} \ar[ru] & \vdots &&&& \\
U_{X_{5}} \ar[r] & *+[F-:<10pt>]{X_{5}} & *+[F-:<10pt>]{X_{10}(C = \emptyset)} \ar[l] &&  \\
}
$$
Note that, for this particular example, the counterfactual features, $X_{j}(C = \emptyset)$, are computed in exactly the same way as the causality-aware training features, $X_{j}^\ast = X_{j} - \hat{\beta}_{{X_j}C} \, C$. Therefore, application of the counterfactual normalization approach to the training set alone would produce the same results as the poor man's causality-aware method, while application of counterfactual normalization to both training and test set features would produce the same results as the causality-aware approach in this example.

\subsection{Classification task experiments}

Even though our analyses were based on the expected MSE metric, and are not applicable to other performance metrics, here we provide empirical evidence that it is still important to deconfound the test set features in classification tasks evaluated with the AUROC metric\footnote{See Chaibub Neto 2020a for analogous results for general performance metrics, and Chaibub Neto 2020b for analogous results for the classification accuracy metric.}.

For the classification task experiments we simulate large datasets according to the model,
$$
\xymatrix@-1.2pc{
& & & *+[F-:<10pt>]{C} \ar[dll]_{\beta_{{X_1}C}} \ar[ddll] \ar[ddddll] \ar[ddr]^{\beta_{YC}} &  \\
U_{X_1} \ar[r] \ar@/_1pc/@{<->}[dd] \ar@/_1.5pc/@{<->}[ddd] & *+[F-:<10pt>]{X_{1}} &&&& \\
& *+[F-:<10pt>]{X_{2}} & & & *+[F-:<10pt>]{Y} \ar[ulll] \ar[lll] \ar[ddlll]^{\beta_{{X_{10}}Y}}  \\
U_{X_{2}} \ar[ru] & \vdots &&&& \\
U_{X_{5}} \ar[r] & *+[F-:<10pt>]{X_{5}} &&&  \\
}
$$
where $C$ and $Y$ are now binary variables, and then apply a selection mechanism in order to obtain subsamples that are affected by selection biases.

The data is generated according to the following model,
\begin{align}
C &\sim \mbox{Bernoulli}(1/2)~, \label{eq:C.class.model} \\
Y &\sim \mbox{Bernoulli}(p)~, \label{eq:Y.class.model} \\
&\;\;\;\;p = 1/(1 + \exp{\{ -\beta_{YC} \, C \}}) \nonumber \\
X_{j} &= \beta_{{X_j}Y} \, Y + \beta_{{X_j}{C}} \, C + U_{X_j}~, \label{eq:X.class.model}
\end{align}
for $j = 1, \ldots, 5$, where the correlated error terms $U_{X_1}$, \ldots, $U_{X_5}$ are simulated as described before.

Since $C$ and $Y$ are binary we have that the support of the joint distribution $P(C, Y)$ contains only the values, $\{C = 0, Y = 0\}$, $\{C = 0, Y = 1\}$, $\{C = 1, Y = 0\}$, and $\{C = 1, Y = 1\}$ and we generate selection biases by selectively subsampling the large dataset using different probabilities $p_{ij} = P(C = i, Y = j)$ for selecting a pair $\{C = i, Y = j\}$. For instance, in order to generate a subsample with strong positive association between $C$ and $Y$ we can select samples from the large dataset according to the probabilities, $p_{00} = 0.4$, $p_{01} = 0.10$, $p_{10} = 0.10$, and $p_{11} = 0.4$. Similarly, in order to generate a subsample with strong negative association, we can select samples with, for example, $p_{00} = 0.1$, $p_{01} = 0.40$, $p_{10} = 0.40$, and $p_{11} = 0.1$.

Note that by subsampling from the large dataset according to $p_{00}$, $p_{01}$, $p_{10}$, and $p_{11}$ we enforce that our samples of the $\{C, Y\}$ variables are distributed according to a bivariate Bernoulli distribution (Dai, Ding, and Wahba 2013) where,
\begin{equation*}
Cov(C, Y) = p_{11} \, p_{00} \, - \, p_{01} \, p_{10}~.
\end{equation*}
Furthermore, because the marginal distributions for $C$ and $Y$ correspond, respectively, to Bernoulli distributions with probability of success equal to $p_{10} + p_{11}$ and $p_{01} + p_{11}$ (Dai, Ding, and Wahba 2013), we have that,
\begin{align*}
Var(C) &= (p_{01} + p_{11})(1 - p_{01} - p_{11})~, \\
Var(Y) &= (p_{10} + p_{11})(1 - p_{10} - p_{11})~.
\end{align*}

As before, we perform two experiments. The first, where $Var(Y_{ts})$ is constant across the test sets, and the second, where it varies. Our experiments were based on 1000 simulations. For each simulation replication we:
\begin{enumerate}
\item Sample the simulation parameters $\beta_{{X_j}Y}$, $\beta_{{X_j}{C}}$, and $\beta_{{Y}{C}}$ from a $U(-1, 1)$ distribution, and $\rho$ from a $U(-0.5, 0.5)$ distribution.
\item Generate a large dataset ($n = 10,000$) according to equations (\ref{eq:C.class.model}), (\ref{eq:Y.class.model}), and (\ref{eq:X.class.model}), using the simulation parameter values sampled in step 1, and then generate the training set by sampling from the large population with probabilities $p_{00} = 0.45$, $p_{01} = 0.05$, $p_{10} = 0.05$, and $p_{11} = 0.45$.
\item Generate 9 distinct test sets (each containing $n = 1,000$ samples). Each test set was generated as follows:
\begin{enumerate}
\item Generate a large dataset ($n = 10,000$) according to equations (\ref{eq:C.class.model}), (\ref{eq:Y.class.model}), and (\ref{eq:X.class.model}) using the simulation parameter values sampled in step 1.
\item Obtain the test set ($n=1,000$) by sampling from the large dataset according to the probabilities presented in Table \ref{tab:parameter.values.1} for the first experiment, and from Table \ref{tab:parameter.values.2} for the second experiment.
\end{enumerate}
\item For the causality-aware approach we: adjust the training set, and each of the 9 distinct test sets; fit a logistic regression model on the adjusted training set data; use the same trained model to predict on the 9 adjusted test sets; and evaluate the test set performances using AUROC.

\item For the ``poor man's" version of the causality-aware approach we: only adjust the training data; fit a logistic regression model to the deconfounded training data; use the trained model to predict on the 9 unadjusted test sets; and evaluate the test set performances using AUROC.

\item For the matching approach we: match the training data alone; fit a logistic regression model to the matched training data; use the trained model to predict on the 9 unadjusted test sets; and evaluate the test set performances using AUROC.

\item For the ``no adjustment" approach we: fit a logistic regression model to the confounded training data; use the trained model to predict on the 9 unadjusted test sets; and evaluate the test set performances using AUROC.
\end{enumerate}

\begin{table*}[!h]
{\scriptsize
\begin{center}
\setlength\tabcolsep{3.0pt}
\begin{tabular}{c|c|ccccccccc}
\hline
 & train. & \multicolumn{9}{c}{test sets} \\
 & set & test 1 & test 2 & test 3 & test 4 & test 5 & test 6 & test 7 & test 8 & test 9 \\
\hline
$p_{11}$ & 0.45 & 0.45 & 0.40 & 0.35 & 0.30 & 0.25 & 0.20 & 0.15 & 0.10 & 0.05 \\
$p_{10}$ & 0.05 & 0.05 & 0.10 & 0.15 & 0.20 & 0.25 & 0.30 & 0.35 & 0.40 & 0.45 \\
$p_{01}$ & 0.05 & 0.05 & 0.10 & 0.15 & 0.20 & 0.25 & 0.30 & 0.35 & 0.40 & 0.45 \\
$p_{00}$ & 0.45 & 0.45 & 0.40 & 0.35 & 0.30 & 0.25 & 0.20 & 0.15 & 0.10 & 0.05 \\
\hline
$Var(C)$ & 0.25  & 0.25 & 0.25 & 0.25 & 0.25 & 0.25 & 0.25 & 0.25 & 0.25 & 0.25 \\
$Var(Y)$ & 0.25  & 0.25 & 0.25 & 0.25 & 0.25 & 0.25 & 0.25 & 0.25 & 0.25 & 0.25 \\
$Cov(C,Y)$ & 0.20 & 0.20 & 0.15 & 0.10 & 0.05 & 0.00 & -0.05 & -0.10 & -0.15 & -0.20 \\
$Cor(C,Y)$ & 0.80  & 0.80 & 0.60 & 0.40 & 0.20 & 0.00 & -0.20 & -0.40 & -0.60 & -0.80 \\
\hline
\end{tabular}
\end{center}}
\vskip -0.2in
\caption{Sampling probabilities for the first experiment (fixed $Var(Y_{ts})$).}
\label{tab:parameter.values.1}
\end{table*}

\begin{table*}[!h]
{\scriptsize
\begin{center}
\setlength\tabcolsep{3.0pt}
\begin{tabular}{c|c|ccccccccc}
\hline
 & train. & \multicolumn{9}{c}{test sets} \\
 & set & test 1 & test 2 & test 3 & test 4 & test 5 & test 6 & test 7 & test 8 & test 9 \\
\hline
$p_{11}$ & 0.45 & 0.45 & 0.40 & 0.35 & 0.30 & 0.25 & 0.20 & 0.15 & 0.10 & 0.05 \\
$p_{10}$ & 0.05 & 0.05 & 0.15 & 0.25 & 0.35 & 0.45 & 0.55 & 0.65 & 0.75 & 0.85 \\
$p_{01}$ & 0.05 & 0.05 & 0.05 & 0.05 & 0.05 & 0.05 & 0.05 & 0.05 & 0.05 & 0.05 \\
$p_{00}$ & 0.45 & 0.45 & 0.40 & 0.35 & 0.30 & 0.25 & 0.20 & 0.15 & 0.10 & 0.05 \\
\hline
$Var(C)$ & 0.25  & 0.25 & 0.2475 & 0.2400 & 0.2275 & 0.2100 & 0.1875 & 0.1600 & 0.1275 & 0.0900 \\
$Var(Y)$ & 0.25  & 0.25 & 0.2475 & 0.2400 & 0.2275 & 0.2100 & 0.1875 & 0.1600 & 0.1275 & 0.0900 \\
$Cov(C,Y)$ & 0.20  & 0.20 & 0.1525 & 0.1100 & 0.0725 & 0.0400 & 0.0125 & -0.0100 & -0.0275 & -0.0400 \\
$Cor(C,Y)$ & 0.80 & 0.80 & 0.6162 & 0.4583 & 0.3187 & 0.1905 & 0.0667 & -0.0625 & -0.2157 & -0.4444 \\
\hline
\end{tabular}
\end{center}}
\vskip -0.2in
\caption{Sampling probabilities for the second experiment (varying $Var(Y_{ts})$).}
\label{tab:parameter.values.2}
\end{table*}

\begin{figure*}[!h]
\centerline{\includegraphics[width=\linewidth]{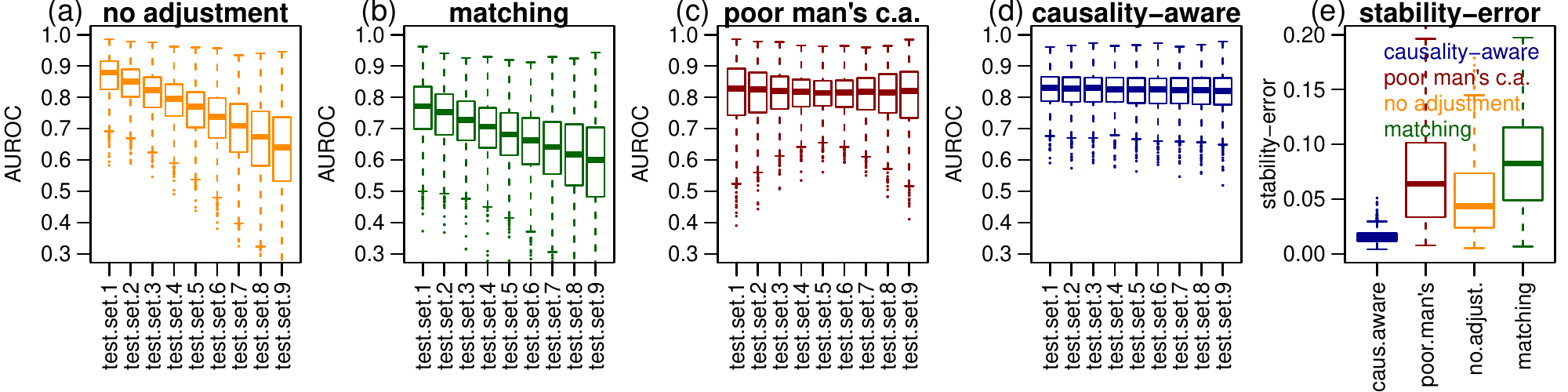}}
\vskip -0.15in
\caption{Classification task synthetic data experiments.}
\label{fig:mpower.aucs.stability.class.1}
\end{figure*}

\begin{figure*}[!h]
\centerline{\includegraphics[width=\linewidth]{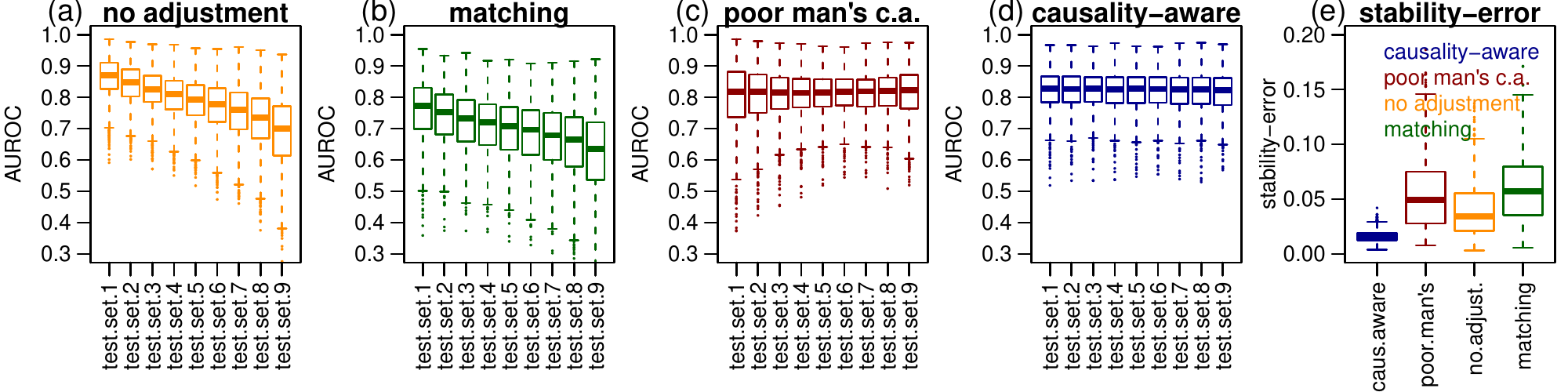}}
\vskip -0.15in
\caption{Classification task synthetic data experiments.}
\label{fig:mpower.aucs.stability.class.2}
\end{figure*}

Note that, as before, only $Cov(Y, C)$, $Var(C)$, and $Var(Y)$ are allowed to vary across the training and test sets, while we still use the same values of the $\beta_{{X_j}Y}$, $\beta_{{X_j}{C}}$, $\beta_{{Y}{C}}$, and $\rho$ parameters in the simulation of the training and test sets (so that the feature model $P(\bfX \mid C, Y)$ is still stable across the datasets). Observe, as well, that we again have that the first test set is generated using the same values of $Cov(Y, C)$, $Var(C)$, and $Var(Y)$ as the training set, illustrating the case where the training and test sets are independent and identically distributed.

Figure \ref{fig:mpower.aucs.stability.class.1} reports the results from experiment 1 (which are also reported in Figure \ref{fig:synthetic.data.stability}b in the main text) where $Var(Y_{ts})$ is kept constant across the test sets, while Figure \ref{fig:mpower.aucs.stability.class.2} reports the results from experiment 2, where $Var(Y_{ts})$ varies across the test sets. Both experiments illustrate the better stability of the causality-aware approach.

Quite interesting these results based on the AUROC metric are less impacted by shifts in $Var(Y_{ts})$ than the regression task results. This observation illustrates the fact that stability results can also depend on the type of metric used the evaluate the predictive performance.


\section{Additional details - real data experiments}

We also validated our stability results using the mPower data (Bot et al. 2016), a mHealth study in Parkinson's disease. As pointed by Chaibub Neto et al (2019), age is an important confounder in this dataset, and our goal is to build classifiers of disease status (i.e., PD vs non-PD) using feature representations learned by a deep model trained on the raw accelerometer data. (For our analyses we discretize age into three levels, ``young age", ``middle-age", and ``senior-age".) In our experiments we adopted the features learned by a top performing deep learning team (``ethz-dreamers") in the Parkinson's Disease Digital Biomarker (PDDB) Dream Challenge (Sieberts et al 2020). We selected the ``ethz-dreamers" submission\footnote{Described in the wiki \texttt{https://www.synapse.org /\#!Synapse:syn10922704/wiki/471154}} as it represents the best deep learning-based submission to the Challenge. (This team got 2nd place in the Challenge. An ensemble method secured the 1st place).

\begin{figure}[!h]
\centerline{\includegraphics[width=2.8in]{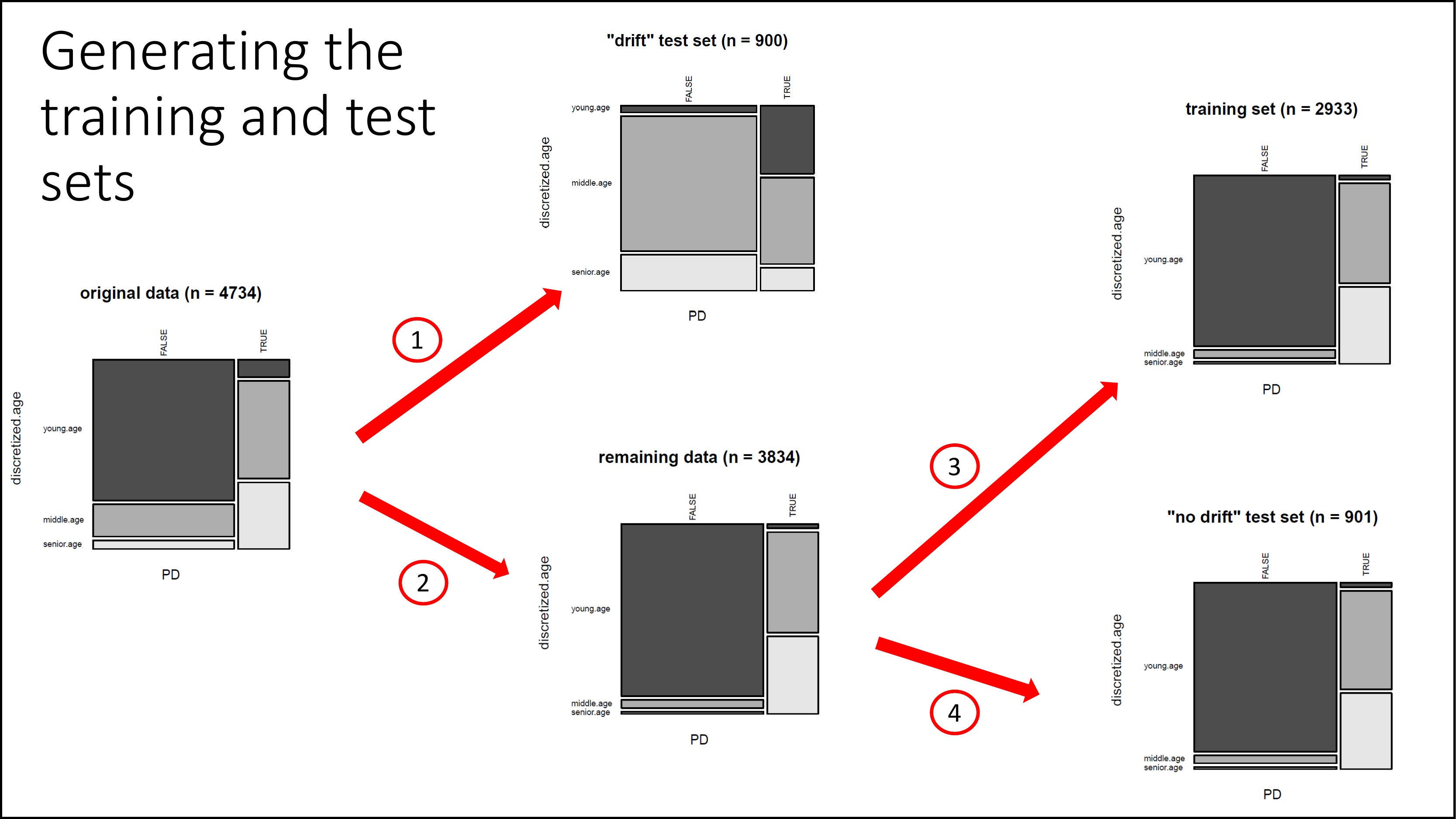}}
\vskip -0.1in
\caption{Data splitting procedure.}
\label{fig:mpower.data.split}
\end{figure}

In order to generate test sets that are shifted with respect to the training data (denoted ``drift test set"), as well as, test sets that are not shifted (denoted ``no drift test set"), we adopted the following strategy (illustrated in Figure \ref{fig:mpower.data.split}). First, we split the original data into a ``drift test set" and a ``remaining data" set by taking a subsample ($n = 900$) of the original data where the association between the (discretized) age confounder and the disease labels is flipped relative to the original data (as described by arrows 1 and 2 in Figure \ref{fig:mpower.data.split}). Second, we split the remaining data into a training set ($n=2933$ as shown in arrow 3) and a identically distributed test set, the ``no drift test set" ($n = 901$ as shown by arrow 4 in Figure \ref{fig:mpower.data.split}). For our analyses we perform 100 distinct data-splits using this procedure.

In these experiments we compared 5 distinct adjustment approaches:
\begin{enumerate}
\item The causality-aware approach, where we adjusted both the training and test set features.
\item The poor man's version of the causality-aware approach, where only the training set features are adjusted.
\item The matching approach (applied to the training set alone), where we remove the association between the (discretized) age confounder, $C$, and the PD status labels, $Y$, by randomly selecting a perfectly balanced subsample of the training set.
\item The approximate inverse probability weighting approach (applied to the training set alone), where we remove the association between $C$ and $Y$ by oversampling from the training data according to weights derived from standard propensity scores (estimated with logistic regression) in order to generated a perfectly balanced augmented training set.
\item The ``no adjustment" approach, where no adjustment is applied to the training or test set.
\end{enumerate}

\begin{figure}[!h]
\centerline{\includegraphics[width=2.0in]{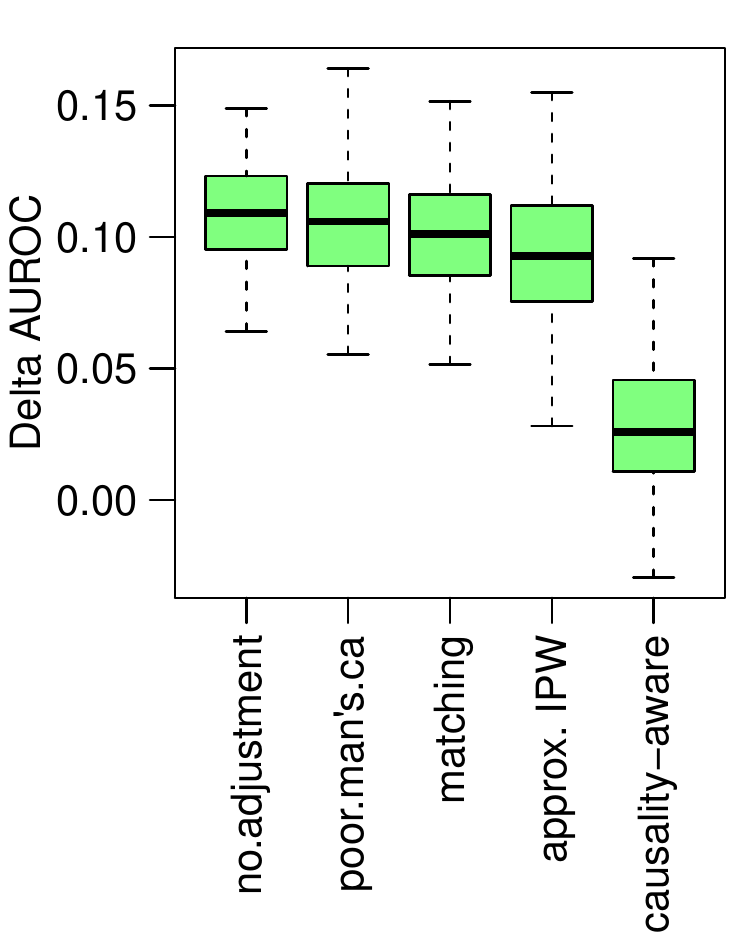}}
\vskip -0.1in
\caption{Drop in performance due to dataset shift. Delta AUROC is computed as the difference in the AUROC from the ``no shift" test set minus the AUROC from the ``shifted" test set.}
\label{fig:mpower.aucs.gap.stability}
\end{figure}

The results are presented in Figure \ref{fig:mpower.aucs.stability} in Section 5, and illustrate the better stability (i.e., the considerably smaller gap between the performances in the ``shift" vs ``no shift" test sets). Finally, observe that, in this example, we have that the confounding adjustment reduces the predictive performance because the spurious associations generated by the age confounder increase the total association between the features and the outcome. (This is illustrated by the drop in performance observed across all adjustment methods relative to the ``no adjustment" approach in the ``no shift" test set.) However, observe, as well, that the ``no adjustment" approach generates the most unstable results as it shows the largest gap in performance between the ``shift" and ``no shift" test sets (as illustrated in Figure \ref{fig:mpower.aucs.gap.stability}).

\end{document}